\documentclass[runningheads]{llncs}
\usepackage{graphicx}
\usepackage{pslatex}
\usepackage{apacite}

\usepackage{multirow}
\usepackage{color}
\usepackage{graphicx}
\usepackage{booktabs} 
\usepackage{pslatex}
\usepackage{apacite}
\usepackage{amssymb}
\usepackage{amsmath}
\usepackage{multicol}
\usepackage{natbib}
\usepackage{array}
\usepackage{subfig}

\newcolumntype{L}[1]{>{\raggedright\arraybackslash}m{#1}}
\newcolumntype{C}[1]{>{\centering\arraybackslash}m{#1}}
\newcolumntype{R}[1]{>{\raggedleft\arraybackslash}m{#1}}
\begin{document}
\title{DIM: Dynamic Integration of Multimodal Entity Linking with Large Language Model}
%

\author{
  Shezheng Song\textsuperscript{1}, Shasha Li\textsuperscript{1}, Jie Yu\textsuperscript{1}, Shan Zhao\textsuperscript{2}, Xiaopeng Li\textsuperscript{1}, Jun Ma\textsuperscript{1}, \\
  Xiaodong Liu\textsuperscript{1}, Zhuo Li\textsuperscript{1}, Xiaoguang Mao\textsuperscript{1}\\
}

\institute{
National University of Defense Technology \and
Hefei University of Technology\\ 
\email{\{betterszsong\}@gmail.com}}

%
%
%
\maketitle              
\begin{abstract}
Our study delves into Multimodal Entity Linking, aligning the mention in multimodal information with entities in knowledge base. Existing methods are still facing challenges like ambiguous entity representations and limited image information utilization. 
Thus, we propose dynamic entity extraction using ChatGPT, which dynamically extracts entities and enhances datasets. We also propose a method: Dynamically Integrate Multimodal information with knowledge base (DIM), employing the capability of the Large Language Model (LLM) for visual understanding. The LLM, such as BLIP-2, extracts information relevant to entities in the image, which can facilitate improved extraction of entity features and linking them with the dynamic entity representations provided by ChatGPT.
The experiments demonstrate that our proposed DIM method outperforms the majority of existing methods on the three original datasets, and achieves state-of-the-art (SOTA) on the dynamically enhanced datasets (Wiki+, Rich+, Diverse+).
For reproducibility, our code and collected datasets are released on \url{https://github.com/season1blue/DIM}.

——
\end{abstract}

\section{Introduction}
    
    Multimodal entity linking~\citep{cel} involves linking mentioned entity (i.e. mention) in natural language texts to their corresponding entity~\citep{zhao2021dynamic} in a knowledge base~\citep{fu2020kbsurvey}. These entities could be individuals, locations, organizations in the real world, or specific entities within a knowledge base. 
    MEL could help computers better understand user semantics, integrate various information sources, resolve ambiguities, and enhance user experience. It plays an important role in search engines \citep{wu2024v}, recommendation systems \citep{zhou2023mmrec}, information retrieval\citep{ma2023adaptive}, and knowledge graph construction \citep{ma2023using, zhao2021dynamic}. Through MEL, systems can provide more relevant search results, more personalized recommendations, more comprehensive information retrieval, richer knowledge graphs, and smarter conversation and text processing capabilities.

    Human cognition~\citep{hu2021can, hutchins2020distributed} can be expressed through various modalities of information carriers, such as text or images.
    Building a connection between natural language and structured knowledge contributes to the unity of human cognition and the knowledge base, enabling artificial intelligence to comprehend human society better.
    Specifically, multimodal entity linking aids in a deeper understanding of the semantics of text. By linking entities in the expression to a knowledge base, systems can acquire additional information about these entities, elevating the comprehension of textual meanings. 
    Besides fostering the unity of semantic expression, multimodal entity linking also assists in enriching the information in knowledge bases. Through entity linking, new relationships and attributes can be added to the knowledge base, thereby enhancing its completeness and accuracy.
    \begin{figure*}[tb]
        \centering
        \includegraphics[width=.8\textwidth]{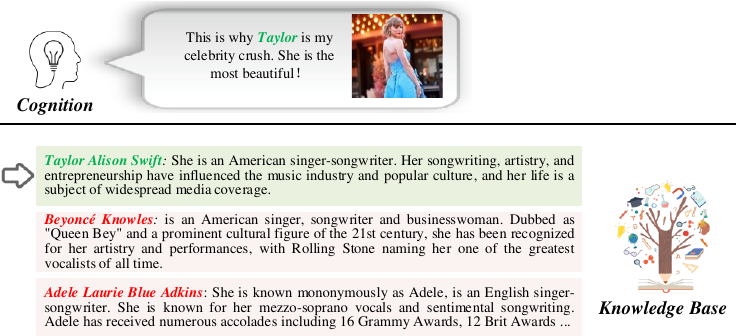}
        \caption{The process of integrating human cognition with information in the knowledge base. When someone says ``This is why Taylor ....", MEL aims to link the mentioned 'Taylor' to 'Taylor Alison Swift' in knowledge base to facilitate further understanding of user semantics.}
        \label{fig:intro}
    \end{figure*}

    In recent years, the multimodal entity linking task~\citep{zhao2021dynamic2} has attracted increasing attention from researchers. \citet{baseline_mel} adopts pretrained models such as BERT and constructs a multi-level multimodal information understanding hierarchy to accomplish entity linking. Both the \citet{wikidiverse} and \citet{dwe} enhance the entity linking task by proposing new datasets and fine-grained features.
    However, existing methods face two main challenges:
    \textbf{(1)} Researchers have not adequately considered that \textbf{ambiguous entity representations}~\citep{he2013learning} in datasets may not effectively represent entities in the knowledge base. Typically, textual descriptions are used as representations of entities in the knowledge base, and if the computed features are deemed similar to entity representations, the entity linking task is considered complete. However, in this process, the linking process may fail due to the misalignment between ambiguous entity representations and the semantics of knowledge base entities even with well-learned features. Existing methods overly focus on enhancing the feature learning aspect, overlooking the ambiguity in entity representations~\citep{ji2022win}.
    \textbf{(2)} Existing work still exhibits \textbf{limited understanding of images}. As a crucial supplementary source to text, image information deserves more attention. Methods like GHMFC~\citep{baseline_mel} and MMEL~\citep{MMEL} are trained by encoding images to enhance the linking ability of image information to entity representations. However, the impact of image information on recognizing entity identities is limited. While image encoder can extract racial features, such as ``white female, blue dress", it struggles to identify the identity information due to a lack of comprehensive understanding of image.

    In light of the aforementioned challenges, we propose the following optimization methods:
    \textbf{(1)} Addressing the issue of ambiguous entity representations that hinder the unity of human cognition and knowledge base, we leverage the rapid learning capabilities of ChatGPT on the knowledge base and dynamically extract representations of entities. To facilitate subsequent experiments by researchers, we organize the dynamically extracted entities and enhance three datasets (Wikimel, Richpedia, and Wikidiverse) individually. We designate these enhanced datasets as Wiki+, Rich+, and Diverse+.
    \textbf{(2)} Addressing the inadequacy in information utilization by existing models, we propose a method to Dynamically Integrate Multimodal information (DIM). Besides, we conduct experiments on both the original datasets(Wikimel, Richpedia, and Wikidiverse) and the enhanced datasets(Wiki+, Rich+, Diverse+) to validate the effectiveness of our datasets. Specifically, we leverage the understanding capabilities of the Large Language Model~\citep{li2022blip, blip2} on images and design prompts to extract information from images, such as obtaining image titles or inquiring about the identities of individuals in the images. Applying LLM to the human images as shown in Figure \ref{fig:intro} will extract information such as ``Taylor Swift on wedding", which contributes to linking the entity.
    
    
    In summary, our innovations are outlined as follows:
    \begin{itemize}
        \item We propose a dynamic method for collecting entity representations from the knowledge base, and release the collected enhanced datasets Wiki+, Rich+, and Diverse+.
        \item We introduce a dynamic linking method, DIM, to connect human cognition with the knowledge base. DIM utilizes BLIP-2 for better extraction of entity-related features from images and links them with the dynamic entity representations provided by ChatGPT.
        \item Extensive experiments are conducted. The DIM method not only outperforms most methods on the original Wikimel, Richpedia, and Wikidiverse datasets but also achieves state-of-the-art performance on Wiki+, Rich+, and Diverse+.
    \end{itemize}

\section{Dynamic Building Entity Representation}
    %

    Multimodal entity linking(MEL) significantly influences and facilitates a profound understanding and cognition of information for humans. MEL serves as a crucial means to unify human cognition with structured knowledge repositories:
    (1) \textbf{Assurance of semantic consistency}: MEL ensures semantic consistency by aligning entities mentioned in the cognition with knowledge base. It helps in eliminating ambiguity and ensuring that the interpretation of specific entities remains clear despite contextual variations.
    (2) \textbf{Enhancement of cognitive information}: MEL offers individuals a richer and deeper cognitive experience. By associating entities with background knowledge, individuals can comprehensively grasp the meaning of entities, thereby elevating their cognitive awareness of information.
    (3) \textbf{Integrated knowledge acquisition}: This contributes to breaking down information silos, enabling people to easily cross different domains, texts, and knowledge sources to acquire information, promoting an overall improvement in cognitive levels.
    
    \subsection{Existing Entity Representation}
        In our investigation and study, we analyze the existing Entity Linking datasets and their methods of entity representation:

        \begin{itemize}
            \item \textbf{Wikimel} and \textbf{Richpedia}~\citep{baseline_dataset} employ concise attributes from Wikidata. This representation lacks representativeness for entities, as many entities share similar attributes. It is easy to link accurate cognition to the wrong entity incorrectly.
        
            \item \textbf{Wikiperson}~\citep{wikiperson}, similar to Wikimel, uses attributes of individuals as representatives, but in a more simplified manner. In this example, the attribute ``American President" is inadequate as a representative of Joe Biden, given that there have been multiple American Presidents.
            
            \item \textbf{Weibo}~\citep{weibo} utilizes individuals from Weibo as entities, using user-authored personal bios as entity representations. These bios, relying on user-generated content, may contain biases or errors and do not accurately reflect the broader public's understanding of the entity.
        
            \item \textbf{Wikidiverse}~\citep{wikidiverse} uses images collected from Wikipedia as entity representation. However, images can deviate from a person's true appearance due to factors like angles and time, lacking real-time accuracy. 
        
        \end{itemize}

\begin{table}[htbp]
  \centering
  \caption{Examples of entity representations for the entity \textbf{Joe Biden} in different datasets. \textbf{Wiki+} is the dataset with dynamic enhancement.}
    \begin{tabular}{p{11.5cm}}
    \toprule
    \multicolumn{1}{l}{\textbf{Wikimel}} \\
    Sex: male. Birth: 1942, Scranton. Religion: Catholicism. Occupation: lawyer, politician. Spouse: Jill Biden, Neilia Hunter. Languages: English. Alma mater: Archmere Academy, Syracuse University College of Law... \\
    \midrule
    \multicolumn{1}{l}{\textbf{Wikiperson}} \\
    President of the United States \\
    \midrule
    \multicolumn{1}{l}{\textbf{Weibo}} \\
    Husband to @DrBiden, proud father and grandfather. Ready to finish the job for all Americans. \\
    \midrule
    \multicolumn{1}{l}{\textbf{Wikidiverse}} \\
    Joe Biden became the presumptive nominee of the Democratic Party for president in April 2020, and formally accepted the nomination the following August... \\
    \midrule
    \multicolumn{1}{l}{\textbf{Wiki+(Ours)}} \\
    Joe Biden is an American politician who served as the 46th president of the United States. Born on November 20, 1942, in Scranton, Pennsylvania, Biden has had a long and distinguished political career... \\
    \bottomrule
    \end{tabular}%
  \label{tab:er}%
\end{table}%

        In summary, existing multimodal entity linking methods suffer from the limitation that entity representation fails to effectively represent entities. More importantly, these representations are manually collected from Wikipedia or other knowledge bases and can only represent the entity's state at a specific time. Human understanding of entities changes over time and events. For instance, Donald Trump is no longer the President of the United States in 2023. In such cases, rigid and less adaptable entity representations can lead to errors.
        Additionally, when a mentioned entity is not in the dataset, there is no corresponding entity representation, causing potential issues in entity linking. 

        Table \ref{tab:er} shows the diverse representation methods for entities in existing research. (It is worth noting that Weibo dataset~\citep{weibo} is primarily designed for Chinese celebrities and does not involve international figures. For the sake of convenient presentation and visual comparison, following the definition of the Weibo dataset, we manually gathered personal profiles of relevant entities from Twitter. )
        Therefore, the paper proposes a dynamic method for constructing entity representations, leveraging the capabilities of ChatGPT to enhance real-time and scalable entity representations based on the evolving understanding of the world.
    
    \subsection{Leverage Chatgpt to Dynamic Connect}

        
        ChatGPT~\citep{GPT4} is a powerful large model trained on massive amounts of web data and is continuously updated over time. We plan to utilize the interface provided by ChatGPT to inquire about entities, so as to subsequently link with the entities in the implicit knowledge base of ChatGPT. Candidate entities are input into ChatGPT for inquiries using the prompt: ``You are a helpful assistant designed to give a comprehensive introduction about people. Who is this one?" The generated response from ChatGPT is shown in Table \ref{tab:er}.
        For experimental convenience and dataset quality enhancement, we collect the response of ChatGPT to enhance the dataset. In detail, we construct entity representations for 17391, 17804, and 57007 entities from Wikimel, Richpedia, Wikidiverse, respectively. The newly built entity representations better reflect the general public's understanding of entities, align closely with their inherent semantics, and facilitate a unified approach to cognition and knowledge base.  


        Taking the Wikimel~\citep{baseline_dataset} as an example, out of the 17474 entities collected from the dataset, 131 entities did not return any results, and 220 inquiries returned ``Sorry, I cannot provide an introduction to this entity." 
        Besides, ChatGPT provides speculative information based on cultural, regional, or other contextual cues for 462 entities. For example, ``John Abbott is a common English given name and surname," but did not provide specific representation. Additionally, 2997 entities require additional information for verification. For instance, ``It is possible that Edward J. Livernash is a private individual without any notable achievements." Furthermore, 599 entities are speculated to be fictional names, such as ``John McDuffie is a fictional name, so there is no information."
        The specific reasons and their proportions are as shown in Figure \ref{tab:statistics}.
        In summary, out of 17391 entities, 5517 entities were unable to be enhanced through ChatGPT. For these entities, we continue to use the original entity representations.

    \begin{figure*}[tb]
        \centering
        \includegraphics[width=.8\textwidth]{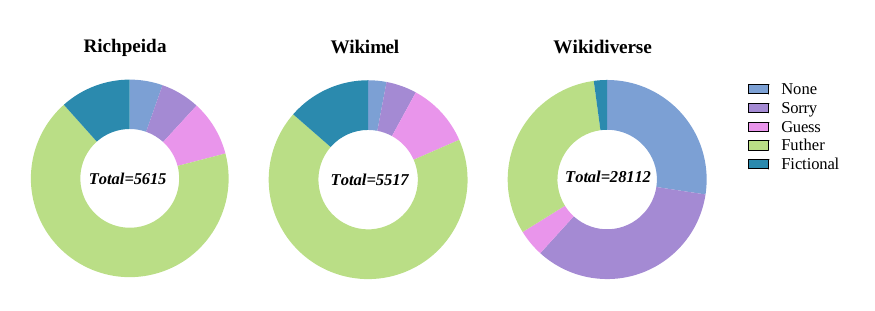}
        \caption{Statistics of enhanced datasets including Richpedia, Wikimel, and Wikidiverse.}
        \label{fig:enhance_statistics}
    \end{figure*}

    \begin{figure*}[tb]
        \centering
        \includegraphics[width=.7\textwidth]{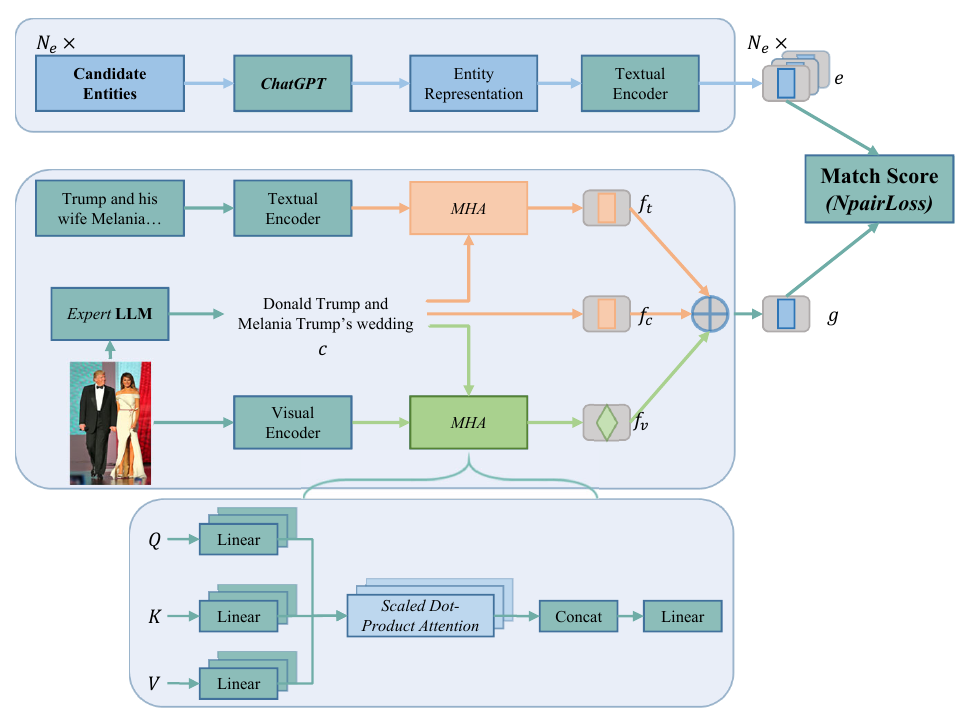}
        \caption{Model overview. Example is an image with mention $m$ \textit{Trump}, text $t$ \textit{Trump and his wife Melania at Wedding}. $c$ is the result of the expert model. Npairloss in contrastive learning is to ensure close distances for same-category samples and distinct distances for different-category samples.}
        \label{fig:model}
    \end{figure*}

\section{Dynamically Integrate Multimodal Information}
    

    To evaluate the effectiveness of our enhanced dataset, we introduce a corresponding baseline, a method to Dynamically Integrate Multimodal information(DIM). DIM was experimented not only on the original Wikimel, Richpedia, and Wikidiverse datasets but also on the enhanced Wiki+, Rich+, and Diverse+ datasets to assess the effectiveness of the enhanced datasets.

    In detail, our DIM employs CLIP~\citep{clip_model} for feature encoding and utilizes BLIP-2~\citep{blip2} as an expert to extract useful information from images, serving as supplementary information for feature extraction by CLIP. This approach was designed to enhance the representation and understanding of entities in our experiments. The performance comparisons across the original and enhanced datasets aimed to highlight the impact of our dataset augmentation on the efficacy of the proposed DIM model.
    
    \subsection{Feature Extractor}
        Given a sentence $x_t$, mention $x_m$ and Wikipedia description of entity $x_e$, we follow CLIP \citep{clip_model} to tokenize it into a sequence of word embeddings. Then the special tokens \textit{startoftext} and \textit{endoftext} are added at the beginning and end positions of word embeddings. 
        As a result, with $N$ sentences and $N_e$ candidate entities, we feed sentence representation  $t \in \mathbb{R}^{N \times d}$,  mention representation $m \in \mathbb{R}^{N \times d}$ and entity representation $e \in \mathbb{R}^{N_e \times d}$ into model.
        Similarly, we feed image into visual encoder of CLIP to get visual feature $v \in \mathbb{R}^{N \times d}$. $d$ is the hidden size of textual and visual features.

    \subsection{Expert Supplementary Information}
        We employ BLIP-2 as the expert for extracting information from images, employing various approaches to process BLIP-2 images:
        1) Image Captioning: We extract corresponding captions $c_1$ for images, such as ``A man and a woman on the red carpet."
        2) Prompt-based Inquiry: We utilize prompts to ask the detailed information $c_2$ about the images, with specific prompt designs such as  ``Question:  Who are the characters in the picture? Answer: "

        We concatenate the image-related information obtained through these two methods to obtain expert information $c=\text{[CLS]}c1\text{[SEP]}c2$, where '[CLS]' and '[SEP]' are special characters used to indicate the beginning and separation of text, respectively. $c$ is fed into the text encoder of CLIP to get expert feature $f_c$.

    \subsection{Architecture}
        We combine the extracted image and text features with the supplementary information provided by the expert, concatenating them to form the final feature representation. To manage the expression of features, we employ multi-head attention for selection. 
        Through multi-head attention, $c$ will interact separately with text feature $t$ and image feature $v$ to extract useful information and control noise.
        \begin{align}
            f_t &= softmax ( \frac{(W^Q f_c)^T (W^K t)}{\sqrt{d}}) (W^V t) \\
            f_v &= softmax ( \frac{(W^Q f_c)^T (W^K v)}{\sqrt{d}}) (W^V v) 
        \end{align}
        where $W^Q \in \mathbb{R}^{d\times d_q}, W^K \in \mathbb{R}^{d\times d_k}, W^V \in \mathbb{R}^{d\times d_v}$ are randomly initialized projection matrices. We set $d_q=d_k=d_v=d/h$. $h$ is the number of heads of attention layer. 
    
        On one hand, text information, image information, and expert features are fused to form the fused feature $g$.
        \begin{equation}
            g = f_v + f_c + f_t
        \end{equation}
        
        On the other hand, the textual representations of $N_e$ candidate entities in the knowledge base also encoded by CLIP's text encoder to obtain $e$. During this process, the limit of text length is imposed to prevent excessively long entity representations. The fused feature $g$ is then matched with the textual features of the $N_e$ entities using cosine similarity, calculating the Match Score. If $g$ is most similar to the correct entity among the $N_e$ entities, it is considered a successful T@1 prediction. Similarly, if the similarity ranks within the top 5, it constitutes a successful T@5 prediction.

    \subsection{Training Loss}
        We utilize Npairloss~\citep{npairloss} as our optimization training objective to enhance the learning process. 
        NpairLoss is applied to networks with multiple outputs, where each output corresponds to a specific task. Its purpose is to optimize the network by maximizing the similarity of sample pairs within the same category, thereby enhancing the effectiveness of multi-task learning.
        Specifically, for each sample, NpairLoss defines the loss by comparing the similarity between positive sample pairs (belonging to the same category) and negative sample pairs (belonging to different categories). It encourages the network to make sample pairs within the same category more similar while ensuring that sample pairs from different categories are more dissimilar.

        The mathematical formulation of NpairLoss $\mathcal{L}_N$ is typically expressed as the sum of losses over each sample pair. The similarity scores of positive sample pairs are maximized, while those of negative sample pairs are minimized. This helps effectively balance the trade-offs between different tasks in multi-task learning, thereby improving the generalization performance of network.
            
        \begin{equation}
            \mathcal{L} =
             \sum_{i=1}^{N} \left( -\frac{\text{sim}(\mathbf{g}_i, \mathbf{p}_i)}{\sum_{j=1}^{K}\text{sim}(\mathbf{g}_i, \mathbf{n}_j)} + \log\left(\sum_{j=1}^{K} e^{\text{sim}(\mathbf{g}_i, \mathbf{n}_j)}\right) \right)
	\end{equation}
        where $g_i$ is the fused feature of the $i$-th sample. $p_i$ is the representation of the positive sample corresponding to the $i$-th sample pair. $n_j$ is the representation of the $j$-th negative sample corresponding to the $i$-th sample pair. $sim(a,b)$ denotes the similarity measure between two representations, typically cosine similarity.

\section{Expreriment}
    DIM method is not only tested on the original dataset, with results presented in Table \ref{tab:result}, but also on the dynamically enhanced dataset we introduced in Table \ref{tab:enhanced}. For comparative analysis, we reproduce several classic baselines on the enhanced dataset, such as BERT~\citep{BERT}, GHMFC~\citep{baseline_mel} and CLIP~\citep{clip_model}.

    \begin{table}[tb]
            \caption{The statistics of datasets. Wiki+, Rich+, Diverse+ represent the enhanced versions of Wikimel, Richpedia, and Wikidiverse datasets using ChatGPT. ``Text" and ``ER" refer to the length of text and entity representation, respectively.}
            \vskip 0.12in
          \centering
            \begin{tabular}{lccccc}
            \toprule
            \textbf{Dataset} & \textbf{Sample} & \textbf{Entity} & \textbf{Mention} & \textbf{Text} & \textbf{ER} \\
            \midrule
            \textbf{Wiki+} & 18880 & 17391 & 25846 & 8.2   & 1318 \\
            \textbf{Rich+} & 17805 & 17804 & 18752 & 13.6  & 1274 \\
            \textbf{Diverse+} & 13765 & 57007 & 16097 & 10.1  & 902 \\
            \bottomrule
            \end{tabular}%
            \label{tab:statistics}%
        \end{table}
    \subsection{Dataset}
        Our experimental dataset comprises authoritative datasets in the entity linking domain, including Richpedia, WikiMEL, Wikidiverse. We conduct a comprehensive analysis of the augmented dataset, and the statistics are presented in Table \ref{tab:statistics}.

    \subsection{Baseline}
    We select several representative methods from the current research community as our baseline:
    \textbf{(1) BLINK} \citep{BLINK} is a two-step entity linking model based on the BERT model.
    \textbf{(2) BERT} \citep{BERT} is a deep learning model based on the attention mechanism and Transformer architecture
    \textbf{(3) ARNN} \citep{ARNN} utilizes the attention-RNN structure to establish the relationship link between entities and input information.
    \textbf{(4) DZMNED} \citep{DZMNED} focuses on utilizing a multimodal attention mechanism to analyze information related to mentions in both images and text.
    \textbf{(5) JMEL} \citep{jmel} leverage fully connected layers to project multimodal features into a shared latent space facilitating the representation of features.
    \textbf{(6) MEL-HI} \citep{MELHI} employs multiple attention mechanisms to focus on different aspects of multimodal information and decrease the effects of noisy images.
    \textbf{(7) HieCoAtt} \citep{HieCoAtt} is a multimodal fusion mechanism, using alternating co-attention and three textual levels (tokens, phrases, and sentences) to calculate relationship.
    \textbf{(8) GHMFC} \citep{baseline_mel} takes the gated multimodal fusion and novel attention mechanism to link entities in knowledge base.
    \textbf{(9) MMEL} \citep{MMEL} is a joint feature extraction module to learn the representations of context and entity candidates, from both the visual and textual perspectives. 
    \textbf{(10) CLIP-text} \citep{clip_model} only uses textual information and focuses on the ability to build textual relationships between text and entity.
     \textbf{(11) CLIP} \citep{clip_model} take both textual and visual features into consideration. The model concatenates multimodal features and calculates the similarity between fused features and entities.

\begin{table*}[ptb]
\renewcommand\arraystretch{1.2}
\caption{Experimental results on the dataset before dynamic enhancement. T@1 denotes Top-1 accuracy, meaning the feature's similarity to the target entity ranks first among multiple candidate entities. Similarly, T@5 signifies the similarity ranking within the top 5. Calculations are conducted among 100 candidate entities.}
\vskip 0.12in
  \centering
    \begin{tabular}{c|cccc|cccc|cccc}
    \toprule
    \multicolumn{1}{c|}{\multirow{2}[2]{*}{\textbf{Models}}} & \multicolumn{4}{c|}{\textbf{Richpedia}} & \multicolumn{4}{c|}{\textbf{Wikimel}} & \multicolumn{4}{c}{\textbf{Wikidiverse}} \\
          & T@1 & T@5 & T@10 & T@20 & T@1 & T@5 & T@10 & T@20 & T@1 & T@5 & T@10 & T@20 \\
    \midrule
    BLINK & 30.8  & 38.8  & 44.5  & 53.6  & 30.8  & 44.6  & 56.7  & 66.4  & -     & 71.2  & -     & - \\
    DZMNED & 29.5  & 41.6  & 45.8  & 55.2  & 30.9  & 50.7  & 56.9  & 65.1  & -     & 39.1  & -     & - \\
    JMEL  & 29.6  & 42.3  & 46.6  & 54.1  & 31.3  & 49.4  & 57.9  & 64.8  & 21.9  & 54.5  & 69.9  & 76.3  \\
    BERT  & 31.6  & 42.0  & 47.6  & 57.3  & 31.7  & 48.8  & 57.8  & 70.3  & 22.2  & 53.8  & 69.8  & 82.8  \\
    ARNN  & 31.2  & 39.3  & 45.9  & 54.5  & 32.0  & 45.8  & 56.6  & 65.0  & 22.4  & 50.5  & 68.4  & 76.6  \\
    MEL-HI & 34.9  & 43.1  & 50.6  & 58.4  & 38.7  & 55.1  & 65.2  & 75.7  & 27.1  & 60.7  & 78.7  & 89.2  \\
    HieCoAtt & 37.2  & 46.8  & 54.2  & 62.4  & 40.5  & 57.6  & 69.6  & 78.6  & 28.4  & 63.5  & 84.0  & 92.6  \\
    GHMFC & 38.7  & 50.9  & 58.5  & 66.7  & 43.6  & 64.0  & 74.4  & 85.8  & -     & -     & -     & - \\
    MMEL  & -     & -     & -     & -     & 71.5  & 91.7  & 96.3  & 98.0  & -     & -     & -     & - \\
    CLIP  & 60.4  & 96.1  & 98.3  & 99.2  & 36.1  & 81.3  & 92.8  & 98.3  & 42.4  & 80.5  & 91.7  & 96.6  \\
    \textbf{DIM}  &  \textbf{66.1 } & \textbf{97.5 } & \textbf{98.9 } & \textbf{99.6 } & \textbf{64.6 } & \textbf{97.6 } & \textbf{99.1 } & \textbf{99.8 } & \textbf{52.7 } & \textbf{94.5 } & \textbf{98.4 } & \textbf{99.5 } \\
    \bottomrule
    \end{tabular}%
  \label{tab:result}%
\end{table*}%
    
    \subsection{Metrics}
        For metric evaluation, we adopted the T@1, 5, 10, and 20 metrics as employed in GHMFC~\citep{baseline_mel}. These metrics represent the ranking of the similarity scores for candidate entities, where the linked entity's similarity score is within the top 1, 5, 10, and 20 positions, respectively. Following the previous approach~\citep{baseline_mel, MMEL}, calculations are conducted among 100 candidate entities.

        To be specific, following the definition in DWE~\citep{dwe}, the formula is as follows:
        \begin{equation}
            \footnotesize
                Acc_{top\text{-}k} = \frac{1}{N}\sum_{i=1}^{N}\eta \{ I(cos(g, gt), cos(g, C_e)) \leq k \}
        \end{equation}
        where $N$ represents the total number of samples, and $\eta$ is the indicator function. When the receiving condition is satisfied, $\eta$ is set to 1, and 0 otherwise. $gt$ is ground truth entity while $C_e$ is a set of candidate entities. $cos$ means cosine similarity function. $I$ is a function to calculate the rank of similarity between joint feature $g$ and ground truth $gt$ among a set of candidate entities $C_e$.

    \subsection{Implement Details}
        Following previous work\citep{baseline_dataset, wikidiverse, dwe}, we select 100 potential entities as candidates. In the Wiki+, Rich+, and Diverse+ datasets, we utilize fuzzy matching \footnote{https://github.com/seatgeek/fuzzywuzzy} technology to recognize candidate entities that resemble the particular mentions.
    
        Our experiments are conducted on RTX 3090 using PyTorch 2.0. ChatGPT is based on the GPT-3.5-turbo version. The version of CLIP employed is Vit-base-patch16-224-in21. The training consisted of 300 epochs, with both image and text hidden layer dimensions set to 512, and the output layer dimension set to 512. We utilized the AdamW optimizer with a learning rate of 5e-5, and the batch size was set to 64.

    \subsection{Experiment on Original Dataset}
    
        We conduct experiments with the DIM method on the datasets before dynamic enhancement to validate its effectiveness. 
        As shown in Table \ref{tab:result}, the performance of the DIM method surpasses the majority of existing models on three public datasets. DIM demonstrates T@1 performance improvements of 5.7\%, 28.5\%, and 10.3\% on the Richpedia, Wikimel, and Wikidiverse datasets, respectively. This indicates that the DIM method is more effective in capturing information related to entity identity in image data.
    
    \subsection{Experiment on Enhanced Dataset}
\begin{table*}[tb]
\caption{Experimental results on the dataset with dynamic enhancement.}
\renewcommand\arraystretch{1.2}
\vskip 0.12in
  \centering
    \begin{tabular}{c|cccc|cccc|cccc}
    \toprule
    \multicolumn{1}{c|}{\multirow{2}[2]{*}{\textbf{Models}}} & \multicolumn{4}{c|}{\textbf{Richpedia}} & \multicolumn{4}{c|}{\textbf{WikiMEL}} & \multicolumn{4}{c}{\textbf{Wikidiverse}} \\
          & T@1 & T@5   & T@10  & T@20  & T@1 & T@5   & T@10  & T@20  & T@1 & T@5   & T@10  & T@20 \\
    \midrule
    BERT  & 35.5  & 77.7  & 87.8  & 94.3  & 32.0  & 75.7  & 88.2  & 95.5  & 10.3  & 23.9  & 33.8  & 47.3  \\
    GHMFC & 34.6  & 77.0  & 87.1  & 93.8  & 33.3  & 75.9  & 88.4  & 95.0  & 14.8  & 29.9  & 39.3  & 53.8  \\
    CLIP  & 63.5  & 95.4  & 97.5  & 98.6  & 63.0  & 96.2  & 98.6  & 99.6  & 45.7  & 88.9  & 96.4  & 99.3  \\
    \textbf{DIM}  & \textbf{65.1 } & \textbf{96.6 } & \textbf{98.3 } & \textbf{99.5 } & \textbf{68.1 } & \textbf{98.4 } & \textbf{99.5 } & \textbf{99.9 } & \textbf{53.4 } & \textbf{95.5 } & \textbf{99.2 } & \textbf{99.7 } \\
    \bottomrule
    \end{tabular}%
  \label{tab:enhanced}%
\end{table*}%

        In Table \ref{tab:statistics}, to illustrate the distinctions between the enhanced datasets and the original datasets, we conducted feature statistics on the datasets and a series of experiments. 
        As shown in Table \ref{tab:enhanced}, not only did we experiment with the DIM method on the enhanced datasets, but for comparison, we also replicated several classic baselines, including BERT, CLIP, and GHMFC.
        Our approach outperforms most existing models on the original dataset without dynamic enhancement, showcasing the effectiveness of our proposed DIM method. Furthermore, on our enhanced dataset(Rich+, Wiki+, and Diverse+), our method continues to demonstrate robust performance, validating the effectiveness of our enhancement approach. The improved entity representation by ChatGPT aligns more coherently with entities in the knowledge base, achieving better semantic consistency.

\section{Conclusion}

    Our study on multimodal entity linking introduces an impactful solution to key challenges. We leverage ChatGPT's rapid learning to enhance datasets (Wiki+, Rich+, Diverse+), addressing ambiguous entity representations. 
    Furthermore, the dynamically integrate multimodal information with knowledge base (DIM) method validates efficacy and improves information extraction from images, overcoming existing limitations. These innovations contribute to a deeper understanding of human cognition and knowledge bases, advancing natural language processing and artificial intelligence. 
    Experiments show that our DIM not only outperforms most methods on the original dataset(Wikimel, Richpedia, Wikidiverse) but also achieves optimal performance on the newly enhanced dataset(Wiki+, Rich+, Diverse+).

    The dataset we collect relies on ChatGPT's understanding of the knowledge base and the world. Although this allows for dynamic entity information linking, it can lead to biases or omissions in data collection due to ChatGPT's potential hallucinations or unavailability. We will continue to explore and refine methods for entity data collection based on large models to enhance accuracy and completeness.

\bibliographystyle{apacite}

\bibliography{bib}

\begin{thebibliography}{}

\bibitem [\protect \citeauthoryear {%
Adjali%
, Besan{\c{c}}on%
, Ferret%
, Le~Borgne%
\BCBL {}\ \BBA {} Grau%
}{%
Adjali%
\ \protect \BOthers {.}}{%
{\protect \APACyear {2020}}%
}]{%
jmel}
\APACinsertmetastar {%
jmel}%
\begin{APACrefauthors}%
Adjali, O.%
, Besan{\c{c}}on, R.%
, Ferret, O.%
, Le~Borgne, H.%
\BCBL {}\ \BBA {} Grau, B.%
\end{APACrefauthors}%
\unskip\
\newblock
\APACrefYearMonthDay{2020}{}{}.
\newblock
{\BBOQ}\APACrefatitle {Multimodal entity linking for tweets} {Multimodal entity linking for tweets}.{\BBCQ}
\newblock
\BIn{} \APACrefbtitle {European Conference on Information Retrieval} {European conference on information retrieval}\ (\BPGS\ 463--478).
\PrintBackRefs{\CurrentBib}

\bibitem [\protect \citeauthoryear {%
Devlin%
, Chang%
, Lee%
\BCBL {}\ \BBA {} Toutanova%
}{%
Devlin%
\ \protect \BOthers {.}}{%
{\protect \APACyear {2019}}%
}]{%
BERT}
\APACinsertmetastar {%
BERT}%
\begin{APACrefauthors}%
Devlin, J.%
, Chang, M\BHBI W.%
, Lee, K.%
\BCBL {}\ \BBA {} Toutanova, K.%
\end{APACrefauthors}%
\unskip\
\newblock
\APACrefYearMonthDay{2019}{Jul}{}.
\newblock
{\BBOQ}\APACrefatitle {BERT: Pre-training of Deep Bidirectional Transformers for Language Understanding} {Bert: Pre-training of deep bidirectional transformers for language understanding}.{\BBCQ}
\newblock
\BIn{} \APACrefbtitle {Proceedings of the 2019 Conference of the North.} {Proceedings of the 2019 conference of the north.}
\newblock
\begin{APACrefURL} \url{http://dx.doi.org/10.18653/v1/n19-1423} \end{APACrefURL}
\newblock
\begin{APACrefDOI} \doi{10.18653/v1/n19-1423} \end{APACrefDOI}
\PrintBackRefs{\CurrentBib}

\bibitem [\protect \citeauthoryear {%
Eshel%
\ \protect \BOthers {.}}{%
Eshel%
\ \protect \BOthers {.}}{%
{\protect \APACyear {2017}}%
}]{%
ARNN}
\APACinsertmetastar {%
ARNN}%
\begin{APACrefauthors}%
Eshel, Y.%
, Cohen, N.%
, Radinsky, K.%
, Markovitch, S.%
, Yamada, I.%
\BCBL {}\ \BBA {} Levy, O.%
\end{APACrefauthors}%
\unskip\
\newblock
\APACrefYearMonthDay{2017}{}{}.
\newblock
{\BBOQ}\APACrefatitle {Named entity disambiguation for noisy text} {Named entity disambiguation for noisy text}.{\BBCQ}
\newblock
\APACjournalVolNumPages{arXiv preprint arXiv:1706.09147}{}{}{}.
\PrintBackRefs{\CurrentBib}

\bibitem [\protect \citeauthoryear {%
Fu%
\ \protect \BOthers {.}}{%
Fu%
\ \protect \BOthers {.}}{%
{\protect \APACyear {2020}}%
}]{%
fu2020kbsurvey}
\APACinsertmetastar {%
fu2020kbsurvey}%
\begin{APACrefauthors}%
Fu, B.%
, Qiu, Y.%
, Tang, C.%
, Li, Y.%
, Yu, H.%
\BCBL {}\ \BBA {} Sun, J.%
\end{APACrefauthors}%
\unskip\
\newblock
\APACrefYearMonthDay{2020}{}{}.
\newblock
{\BBOQ}\APACrefatitle {A survey on complex question answering over knowledge base: Recent advances and challenges} {A survey on complex question answering over knowledge base: Recent advances and challenges}.{\BBCQ}
\newblock
\APACjournalVolNumPages{arXiv preprint arXiv:2007.13069}{}{}{}.
\PrintBackRefs{\CurrentBib}

\bibitem [\protect \citeauthoryear {%
He%
\ \protect \BOthers {.}}{%
He%
\ \protect \BOthers {.}}{%
{\protect \APACyear {2013}}%
}]{%
he2013learning}
\APACinsertmetastar {%
he2013learning}%
\begin{APACrefauthors}%
He, Z.%
, Liu, S.%
, Li, M.%
, Zhou, M.%
, Zhang, L.%
\BCBL {}\ \BBA {} Wang, H.%
\end{APACrefauthors}%
\unskip\
\newblock
\APACrefYearMonthDay{2013}{}{}.
\newblock
{\BBOQ}\APACrefatitle {Learning entity representation for entity disambiguation} {Learning entity representation for entity disambiguation}.{\BBCQ}
\newblock
\BIn{} \APACrefbtitle {Proceedings of the 51st Annual Meeting of the Association for Computational Linguistics (Volume 2: Short Papers)} {Proceedings of the 51st annual meeting of the association for computational linguistics (volume 2: Short papers)}\ (\BPGS\ 30--34).
\PrintBackRefs{\CurrentBib}

\bibitem [\protect \citeauthoryear {%
Hu%
, Lu%
, Pan%
, Gong%
\BCBL {}\ \BBA {} Yang%
}{%
Hu%
\ \protect \BOthers {.}}{%
{\protect \APACyear {2021}}%
}]{%
hu2021can}
\APACinsertmetastar {%
hu2021can}%
\begin{APACrefauthors}%
Hu, Q.%
, Lu, Y.%
, Pan, Z.%
, Gong, Y.%
\BCBL {}\ \BBA {} Yang, Z.%
\end{APACrefauthors}%
\unskip\
\newblock
\APACrefYearMonthDay{2021}{}{}.
\newblock
{\BBOQ}\APACrefatitle {Can AI artifacts influence human cognition? The effects of artificial autonomy in intelligent personal assistants} {Can ai artifacts influence human cognition? the effects of artificial autonomy in intelligent personal assistants}.{\BBCQ}
\newblock
\APACjournalVolNumPages{International Journal of Information Management}{56}{}{102250}.
\PrintBackRefs{\CurrentBib}

\bibitem [\protect \citeauthoryear {%
Hutchins%
}{%
Hutchins%
}{%
{\protect \APACyear {2020}}%
}]{%
hutchins2020distributed}
\APACinsertmetastar {%
hutchins2020distributed}%
\begin{APACrefauthors}%
Hutchins, E.%
\end{APACrefauthors}%
\unskip\
\newblock
\APACrefYearMonthDay{2020}{}{}.
\newblock
{\BBOQ}\APACrefatitle {The distributed cognition perspective on human interaction} {The distributed cognition perspective on human interaction}.{\BBCQ}
\newblock
\BIn{} \APACrefbtitle {Roots of human sociality} {Roots of human sociality}\ (\BPGS\ 375--398).
\newblock
\APACaddressPublisher{}{Routledge}.
\PrintBackRefs{\CurrentBib}

\bibitem [\protect \citeauthoryear {%
Ji%
, Li%
, Yu%
, Ma%
\BCBL {}\ \BBA {} Liu%
}{%
Ji%
\ \protect \BOthers {.}}{%
{\protect \APACyear {2022}}%
}]{%
ji2022win}
\APACinsertmetastar {%
ji2022win}%
\begin{APACrefauthors}%
Ji, B.%
, Li, S.%
, Yu, J.%
, Ma, J.%
\BCBL {}\ \BBA {} Liu, H.%
\end{APACrefauthors}%
\unskip\
\newblock
\APACrefYearMonthDay{2022}{}{}.
\newblock
{\BBOQ}\APACrefatitle {Win-Win Cooperation: Bundling Sequence and Span Models for Named Entity Recognition} {Win-win cooperation: Bundling sequence and span models for named entity recognition}.{\BBCQ}
\newblock
\APACjournalVolNumPages{arXiv preprint arXiv:2207.03300}{}{}{}.
\PrintBackRefs{\CurrentBib}

\bibitem [\protect \citeauthoryear {%
Koml{\'o}si%
\ \BBA {} Waldbuesser%
}{%
Koml{\'o}si%
\ \BBA {} Waldbuesser%
}{%
{\protect \APACyear {2015}}%
}]{%
cel}
\APACinsertmetastar {%
cel}%
\begin{APACrefauthors}%
Koml{\'o}si, L\BPBI I.%
\BCBT {}\ \BBA {} Waldbuesser, P.%
\end{APACrefauthors}%
\unskip\
\newblock
\APACrefYearMonthDay{2015}{}{}.
\newblock
{\BBOQ}\APACrefatitle {The cognitive entity generation: Emergent properties in social cognition} {The cognitive entity generation: Emergent properties in social cognition}.{\BBCQ}
\newblock
\BIn{} \APACrefbtitle {2015 6th IEEE International Conference on Cognitive Infocommunications (CogInfoCom)} {2015 6th ieee international conference on cognitive infocommunications (coginfocom)}\ (\BPGS\ 439--442).
\PrintBackRefs{\CurrentBib}

\bibitem [\protect \citeauthoryear {%
{Li}%
, {Li}%
, {Savarese}%
\BCBL {}\ \BBA {} {Hoi}%
}{%
{Li}%
\ \protect \BOthers {.}}{%
{\protect \APACyear {2023}}%
}]{%
blip2}
\APACinsertmetastar {%
blip2}%
\begin{APACrefauthors}%
{Li}, J.%
, {Li}, D.%
, {Savarese}, S.%
\BCBL {}\ \BBA {} {Hoi}, S.%
\end{APACrefauthors}%
\unskip\
\newblock
\APACrefYearMonthDay{2023}{{\APACmonth{01}}}{}.
\newblock
{\BBOQ}\APACrefatitle {{BLIP-2: Bootstrapping Language-Image Pre-training with Frozen Image Encoders and Large Language Models}} {{BLIP-2: Bootstrapping Language-Image Pre-training with Frozen Image Encoders and Large Language Models}}.{\BBCQ}
\newblock
\APACjournalVolNumPages{arXiv e-prints}{}{}{arXiv:2301.12597}.
\newblock
\begin{APACrefDOI} \doi{10.48550/arXiv.2301.12597} \end{APACrefDOI}
\PrintBackRefs{\CurrentBib}

\bibitem [\protect \citeauthoryear {%
Li%
, Li%
, Xiong%
\BCBL {}\ \BBA {} Hoi%
}{%
Li%
\ \protect \BOthers {.}}{%
{\protect \APACyear {2022}}%
}]{%
li2022blip}
\APACinsertmetastar {%
li2022blip}%
\begin{APACrefauthors}%
Li, J.%
, Li, D.%
, Xiong, C.%
\BCBL {}\ \BBA {} Hoi, S.%
\end{APACrefauthors}%
\unskip\
\newblock
\APACrefYearMonthDay{2022}{}{}.
\newblock
{\BBOQ}\APACrefatitle {Blip: Bootstrapping language-image pre-training for unified vision-language understanding and generation} {Blip: Bootstrapping language-image pre-training for unified vision-language understanding and generation}.{\BBCQ}
\newblock
\BIn{} \APACrefbtitle {International Conference on Machine Learning} {International conference on machine learning}\ (\BPGS\ 12888--12900).
\PrintBackRefs{\CurrentBib}

\bibitem [\protect \citeauthoryear {%
Lu%
, Yang%
, Batra%
\BCBL {}\ \BBA {} Parikh%
}{%
Lu%
\ \protect \BOthers {.}}{%
{\protect \APACyear {2016}}%
}]{%
HieCoAtt}
\APACinsertmetastar {%
HieCoAtt}%
\begin{APACrefauthors}%
Lu, J.%
, Yang, J.%
, Batra, D.%
\BCBL {}\ \BBA {} Parikh, D.%
\end{APACrefauthors}%
\unskip\
\newblock
\APACrefYearMonthDay{2016}{Jan}{}.
\newblock
\APACrefbtitle {Hierarchical Question-Image Co-Attention for Visual Question Answering.} {Hierarchical question-image co-attention for visual question answering.}
\PrintBackRefs{\CurrentBib}

\bibitem [\protect \citeauthoryear {%
Ma%
, Chen%
, Zhou%
, Zhao%
\BCBL {}\ \BBA {} Cai%
}{%
Ma%
, Chen%
\BCBL {}\ \protect \BOthers {.}}{%
{\protect \APACyear {2023}}%
}]{%
ma2023using}
\APACinsertmetastar {%
ma2023using}%
\begin{APACrefauthors}%
Ma, W.%
, Chen, Q.%
, Zhou, T.%
, Zhao, S.%
\BCBL {}\ \BBA {} Cai, Z.%
\end{APACrefauthors}%
\unskip\
\newblock
\APACrefYearMonthDay{2023}{}{}.
\newblock
{\BBOQ}\APACrefatitle {Using multimodal contrastive knowledge distillation for video-text retrieval} {Using multimodal contrastive knowledge distillation for video-text retrieval}.{\BBCQ}
\newblock
\APACjournalVolNumPages{IEEE Transactions on Circuits and Systems for Video Technology}{33}{10}{5486--5497}.
\PrintBackRefs{\CurrentBib}

\bibitem [\protect \citeauthoryear {%
Ma%
, Zhou%
\BCBL {}\ \protect \BOthers {.}}{%
Ma%
, Zhou%
\BCBL {}\ \protect \BOthers {.}}{%
{\protect \APACyear {2023}}%
}]{%
ma2023adaptive}
\APACinsertmetastar {%
ma2023adaptive}%
\begin{APACrefauthors}%
Ma, W.%
, Zhou, T.%
, Qin, J.%
, Xiang, X.%
, Tan, Y.%
\BCBL {}\ \BBA {} Cai, Z.%
\end{APACrefauthors}%
\unskip\
\newblock
\APACrefYearMonthDay{2023}{}{}.
\newblock
{\BBOQ}\APACrefatitle {Adaptive multi-feature fusion via cross-entropy normalization for effective image retrieval} {Adaptive multi-feature fusion via cross-entropy normalization for effective image retrieval}.{\BBCQ}
\newblock
\APACjournalVolNumPages{Information Processing \& Management}{60}{1}{103119}.
\PrintBackRefs{\CurrentBib}

\bibitem [\protect \citeauthoryear {%
Moon%
, Neves%
\BCBL {}\ \BBA {} Carvalho%
}{%
Moon%
\ \protect \BOthers {.}}{%
{\protect \APACyear {2018}}%
}]{%
DZMNED}
\APACinsertmetastar {%
DZMNED}%
\begin{APACrefauthors}%
Moon, S.%
, Neves, L.%
\BCBL {}\ \BBA {} Carvalho, V.%
\end{APACrefauthors}%
\unskip\
\newblock
\APACrefYearMonthDay{2018}{}{}.
\newblock
{\BBOQ}\APACrefatitle {Multimodal named entity disambiguation for noisy social media posts} {Multimodal named entity disambiguation for noisy social media posts}.{\BBCQ}
\newblock
\BIn{} \APACrefbtitle {Proceedings of the 56th Annual Meeting of the Association for Computational Linguistics (Volume 1: Long Papers)} {Proceedings of the 56th annual meeting of the association for computational linguistics (volume 1: Long papers)}\ (\BPGS\ 2000--2008).
\PrintBackRefs{\CurrentBib}

\bibitem [\protect \citeauthoryear {%
{OpenAI}%
}{%
{OpenAI}%
}{%
{\protect \APACyear {2023}}%
}]{%
GPT4}
\APACinsertmetastar {%
GPT4}%
\begin{APACrefauthors}%
{OpenAI}.%
\end{APACrefauthors}%
\unskip\
\newblock
\APACrefYearMonthDay{2023}{{\APACmonth{03}}}{}.
\newblock
{\BBOQ}\APACrefatitle {{GPT-4 Technical Report}} {{GPT-4 Technical Report}}.{\BBCQ}
\newblock
\APACjournalVolNumPages{arXiv e-prints}{}{}{arXiv:2303.08774}.
\newblock
\begin{APACrefDOI} \doi{10.48550/arXiv.2303.08774} \end{APACrefDOI}
\PrintBackRefs{\CurrentBib}

\bibitem [\protect \citeauthoryear {%
Radford%
\ \protect \BOthers {.}}{%
Radford%
\ \protect \BOthers {.}}{%
{\protect \APACyear {2021}}%
}]{%
clip_model}
\APACinsertmetastar {%
clip_model}%
\begin{APACrefauthors}%
Radford, A.%
, Kim, J\BPBI W.%
, Hallacy, C.%
, Ramesh, A.%
, Goh, G.%
, Agarwal, S.%
\BDBL {}Sutskever, I.%
\end{APACrefauthors}%
\unskip\
\newblock
\APACrefYearMonthDay{2021}{{\APACmonth{02}}}{}.
\newblock
\APACrefbtitle {Learning {Transferable} {Visual} {Models} {From} {Natural} {Language} {Supervision}.} {Learning {Transferable} {Visual} {Models} {From} {Natural} {Language} {Supervision}.}
\newblock
\APACaddressPublisher{}{arXiv}.
\newblock
\begin{APACrefURL} [{2022-11-16}]\url{http://arxiv.org/abs/2103.00020} \end{APACrefURL}
\newblock
\APACrefnote{arXiv:2103.00020 [cs]}
\PrintBackRefs{\CurrentBib}

\bibitem [\protect \citeauthoryear {%
Sohn%
}{%
Sohn%
}{%
{\protect \APACyear {2016}}%
}]{%
npairloss}
\APACinsertmetastar {%
npairloss}%
\begin{APACrefauthors}%
Sohn, K.%
\end{APACrefauthors}%
\unskip\
\newblock
\APACrefYearMonthDay{2016}{}{}.
\newblock
{\BBOQ}\APACrefatitle {Improved deep metric learning with multi-class n-pair loss objective} {Improved deep metric learning with multi-class n-pair loss objective}.{\BBCQ}
\newblock
\APACjournalVolNumPages{Advances in neural information processing systems}{29}{}{}.
\PrintBackRefs{\CurrentBib}

\bibitem [\protect \citeauthoryear {%
Song%
\ \protect \BOthers {.}}{%
Song%
\ \protect \BOthers {.}}{%
{\protect \APACyear {2023}}%
}]{%
dwe}
\APACinsertmetastar {%
dwe}%
\begin{APACrefauthors}%
Song, S.%
, Zhao, S.%
, Wang, C.%
, Yan, T.%
, Li, S.%
, Mao, X.%
\BCBL {}\ \BBA {} Wang, M.%
\end{APACrefauthors}%
\unskip\
\newblock
\APACrefYearMonthDay{2023}{}{}.
\newblock
\APACrefbtitle {A Dual-way Enhanced Framework from Text Matching Point of View for Multimodal Entity Linking.} {A dual-way enhanced framework from text matching point of view for multimodal entity linking.}
\PrintBackRefs{\CurrentBib}

\bibitem [\protect \citeauthoryear {%
Sun%
}{%
Sun%
}{%
{\protect \APACyear {2022}}%
}]{%
wikiperson}
\APACinsertmetastar {%
wikiperson}%
\begin{APACrefauthors}%
Sun, W.%
\end{APACrefauthors}%
\unskip\
\newblock
\APACrefYearMonthDay{2022}{}{}.
\newblock
{\BBOQ}\APACrefatitle {Visual Named Entity Linking: A New Dataset and A Baseline} {Visual named entity linking: A new dataset and a baseline}.{\BBCQ}
\newblock
\APACjournalVolNumPages{arXiv preprint arXiv:2211.04872}{}{}{}.
\PrintBackRefs{\CurrentBib}

\bibitem [\protect \citeauthoryear {%
Wang%
}{%
Wang%
}{%
{\protect \APACyear {2022}}%
}]{%
wikidiverse}
\APACinsertmetastar {%
wikidiverse}%
\begin{APACrefauthors}%
Wang.%
\end{APACrefauthors}%
\unskip\
\newblock
\APACrefYearMonthDay{2022}{{\APACmonth{04}}}{}.
\newblock
\APACrefbtitle {{WikiDiverse}: {A} {Multimodal} {Entity} {Linking} {Dataset} with {Diversified} {Contextual} {Topics} and {Entity} {Types}.} {{WikiDiverse}: {A} {Multimodal} {Entity} {Linking} {Dataset} with {Diversified} {Contextual} {Topics} and {Entity} {Types}.}
\newblock
\APACaddressPublisher{}{arXiv}.
\newblock
\begin{APACrefURL} [{2022-12-10}]\url{http://arxiv.org/abs/2204.06347} \end{APACrefURL}
\newblock
\APACrefnote{arXiv:2204.06347 [cs]}
\PrintBackRefs{\CurrentBib}

\bibitem [\protect \citeauthoryear {%
P.~Wang%
}{%
P.~Wang%
}{%
{\protect \APACyear {2022}}%
}]{%
baseline_mel}
\APACinsertmetastar {%
baseline_mel}%
\begin{APACrefauthors}%
Wang, P.%
\end{APACrefauthors}%
\unskip\
\newblock
\APACrefYearMonthDay{2022}{{\APACmonth{07}}}{}.
\newblock
{\BBOQ}\APACrefatitle {Multimodal {Entity} {Linking} with {Gated} {Hierarchical} {Fusion} and {Contrastive} {Training}} {Multimodal {Entity} {Linking} with {Gated} {Hierarchical} {Fusion} and {Contrastive} {Training}}.{\BBCQ}
\newblock
\BIn{} \APACrefbtitle {Proceedings of the 45th {International} {ACM} {SIGIR} {Conference} on {Research} and {Development} in {Information} {Retrieval}} {Proceedings of the 45th {International} {ACM} {SIGIR} {Conference} on {Research} and {Development} in {Information} {Retrieval}}\ (\BPGS\ 938--948).
\newblock
\APACaddressPublisher{Madrid Spain}{ACM}.
\newblock
\begin{APACrefURL} [{2022-11-13}]\url{https://dl.acm.org/doi/10.1145/3477495.3531867} \end{APACrefURL}
\newblock
\begin{APACrefDOI} \doi{10.1145/3477495.3531867} \end{APACrefDOI}
\PrintBackRefs{\CurrentBib}

\bibitem [\protect \citeauthoryear {%
L.~Wu%
, Petroni%
, Josifoski%
, Riedel%
\BCBL {}\ \BBA {} Zettlemoyer%
}{%
L.~Wu%
\ \protect \BOthers {.}}{%
{\protect \APACyear {2019}}%
}]{%
BLINK}
\APACinsertmetastar {%
BLINK}%
\begin{APACrefauthors}%
Wu, L.%
, Petroni, F.%
, Josifoski, M.%
, Riedel, S.%
\BCBL {}\ \BBA {} Zettlemoyer, L.%
\end{APACrefauthors}%
\unskip\
\newblock
\APACrefYearMonthDay{2019}{Nov}{}.
\newblock
\APACrefbtitle {Scalable Zero-shot Entity Linking with Dense Entity Retrieval.} {Scalable zero-shot entity linking with dense entity retrieval.}
\PrintBackRefs{\CurrentBib}

\bibitem [\protect \citeauthoryear {%
P.~Wu%
\ \BBA {} Xie%
}{%
P.~Wu%
\ \BBA {} Xie%
}{%
{\protect \APACyear {2024}}%
}]{%
wu2024v}
\APACinsertmetastar {%
wu2024v}%
\begin{APACrefauthors}%
Wu, P.%
\BCBT {}\ \BBA {} Xie, S.%
\end{APACrefauthors}%
\unskip\
\newblock
\APACrefYearMonthDay{2024}{}{}.
\newblock
{\BBOQ}\APACrefatitle {V?: Guided Visual Search as a Core Mechanism in Multimodal LLMs} {V?: Guided visual search as a core mechanism in multimodal llms}.{\BBCQ}
\newblock
\BIn{} \APACrefbtitle {Proceedings of the IEEE/CVF Conference on Computer Vision and Pattern Recognition} {Proceedings of the ieee/cvf conference on computer vision and pattern recognition}\ (\BPGS\ 13084--13094).
\PrintBackRefs{\CurrentBib}

\bibitem [\protect \citeauthoryear {%
Yang%
\ \protect \BOthers {.}}{%
Yang%
\ \protect \BOthers {.}}{%
{\protect \APACyear {2023}}%
}]{%
MMEL}
\APACinsertmetastar {%
MMEL}%
\begin{APACrefauthors}%
Yang, C.%
, He, B.%
, Wu, Y.%
, Xing, C.%
, He, L.%
\BCBL {}\ \BBA {} Ma, C.%
\end{APACrefauthors}%
\unskip\
\newblock
\APACrefYearMonthDay{2023}{}{}.
\newblock
{\BBOQ}\APACrefatitle {MMEL: A Joint Learning Framework for Multi-Mention Entity Linking} {Mmel: A joint learning framework for multi-mention entity linking}.{\BBCQ}
\newblock
\BIn{} \APACrefbtitle {Uncertainty in Artificial Intelligence} {Uncertainty in artificial intelligence}\ (\BPGS\ 2411--2421).
\PrintBackRefs{\CurrentBib}

\bibitem [\protect \citeauthoryear {%
Zhang%
}{%
Zhang%
}{%
{\protect \APACyear {2021}}%
}]{%
weibo}
\APACinsertmetastar {%
weibo}%
\begin{APACrefauthors}%
Zhang, L.%
\end{APACrefauthors}%
\unskip\
\newblock
\APACrefYearMonthDay{2021}{}{}.
\newblock
{\BBOQ}\APACrefatitle {Attention-based multimodal entity linking with high-quality images} {Attention-based multimodal entity linking with high-quality images}.{\BBCQ}
\newblock
\BIn{} \APACrefbtitle {International Conference on Database Systems for Advanced Applications} {International conference on database systems for advanced applications}\ (\BPGS\ 533--548).
\PrintBackRefs{\CurrentBib}

\bibitem [\protect \citeauthoryear {%
Zhang%
, Li%
\BCBL {}\ \BBA {} Yang%
}{%
Zhang%
\ \protect \BOthers {.}}{%
{\protect \APACyear {2021}}%
}]{%
MELHI}
\APACinsertmetastar {%
MELHI}%
\begin{APACrefauthors}%
Zhang, L.%
, Li, Z.%
\BCBL {}\ \BBA {} Yang, Q.%
\end{APACrefauthors}%
\unskip\
\newblock
\APACrefYearMonthDay{2021}{}{}.
\newblock
{\BBOQ}\APACrefatitle {Attention-based multimodal entity linking with high-quality images} {Attention-based multimodal entity linking with high-quality images}.{\BBCQ}
\newblock
\BIn{} \APACrefbtitle {International Conference on Database Systems for Advanced Applications} {International conference on database systems for advanced applications}\ (\BPGS\ 533--548).
\PrintBackRefs{\CurrentBib}

\bibitem [\protect \citeauthoryear {%
Zhao%
, Hu%
, Cai%
, Chen%
\BCBL {}\ \BBA {} Liu%
}{%
Zhao%
, Hu%
, Cai%
, Chen%
\BCBL {}\ \BBA {} Liu%
}{%
{\protect \APACyear {2021}}%
}]{%
zhao2021dynamic2}
\APACinsertmetastar {%
zhao2021dynamic2}%
\begin{APACrefauthors}%
Zhao, S.%
, Hu, M.%
, Cai, Z.%
, Chen, H.%
\BCBL {}\ \BBA {} Liu, F.%
\end{APACrefauthors}%
\unskip\
\newblock
\APACrefYearMonthDay{2021}{}{}.
\newblock
{\BBOQ}\APACrefatitle {Dynamic modeling cross-and self-lattice attention network for Chinese NER} {Dynamic modeling cross-and self-lattice attention network for chinese ner}.{\BBCQ}
\newblock
\BIn{} \APACrefbtitle {Proceedings of the AAAI Conference on Artificial Intelligence} {Proceedings of the aaai conference on artificial intelligence}\ (\BVOL~35, \BPGS\ 14515--14523).
\PrintBackRefs{\CurrentBib}

\bibitem [\protect \citeauthoryear {%
Zhao%
, Hu%
, Cai%
\BCBL {}\ \BBA {} Liu%
}{%
Zhao%
, Hu%
, Cai%
\BCBL {}\ \BBA {} Liu%
}{%
{\protect \APACyear {2021}}%
}]{%
zhao2021dynamic}
\APACinsertmetastar {%
zhao2021dynamic}%
\begin{APACrefauthors}%
Zhao, S.%
, Hu, M.%
, Cai, Z.%
\BCBL {}\ \BBA {} Liu, F.%
\end{APACrefauthors}%
\unskip\
\newblock
\APACrefYearMonthDay{2021}{}{}.
\newblock
{\BBOQ}\APACrefatitle {Dynamic modeling cross-modal interactions in two-phase prediction for entity-relation extraction} {Dynamic modeling cross-modal interactions in two-phase prediction for entity-relation extraction}.{\BBCQ}
\newblock
\APACjournalVolNumPages{IEEE Transactions on Neural Networks and Learning Systems}{}{}{}.
\PrintBackRefs{\CurrentBib}

\bibitem [\protect \citeauthoryear {%
Zhou%
}{%
Zhou%
}{%
{\protect \APACyear {2021}}%
}]{%
baseline_dataset}
\APACinsertmetastar {%
baseline_dataset}%
\begin{APACrefauthors}%
Zhou, X.%
\end{APACrefauthors}%
\unskip\
\newblock
\APACrefYearMonthDay{2021}{}{}.
\newblock
{\BBOQ}\APACrefatitle {Weibo-mel, Wikidata-mel and Richpedia-mel: multimodal entity linking benchmark datasets} {Weibo-mel, wikidata-mel and richpedia-mel: multimodal entity linking benchmark datasets}.{\BBCQ}
\newblock
\BIn{} \APACrefbtitle {Knowledge Graph and Semantic Computing: Knowledge Graph Empowers New Infrastructure Construction: 6th China Conference, CCKS 2021, Guangzhou, China, November 4-7, 2021, Proceedings 6} {Knowledge graph and semantic computing: Knowledge graph empowers new infrastructure construction: 6th china conference, ccks 2021, guangzhou, china, november 4-7, 2021, proceedings 6}\ (\BPGS\ 315--320).
\PrintBackRefs{\CurrentBib}

\bibitem [\protect \citeauthoryear {%
Zhou%
}{%
Zhou%
}{%
{\protect \APACyear {2023}}%
}]{%
zhou2023mmrec}
\APACinsertmetastar {%
zhou2023mmrec}%
\begin{APACrefauthors}%
Zhou, X.%
\end{APACrefauthors}%
\unskip\
\newblock
\APACrefYearMonthDay{2023}{}{}.
\newblock
{\BBOQ}\APACrefatitle {Mmrec: Simplifying multimodal recommendation} {Mmrec: Simplifying multimodal recommendation}.{\BBCQ}
\newblock
\BIn{} \APACrefbtitle {Proceedings of the 5th ACM International Conference on Multimedia in Asia Workshops} {Proceedings of the 5th acm international conference on multimedia in asia workshops}\ (\BPGS\ 1--2).
\PrintBackRefs{\CurrentBib}

\end{thebibliography}

\end{document}